# Incremental-Smoothing-Based ARAIM Protection Levels for Tightly Coupled GNSS/IMU Integration in Urban Canyons

Jiahao Hu, Xiangru Wang, Weisong Wen, Qian Meng, Jiale Wang, and Shizhuang Wang

***Abstract*—Protection levels (PLs) bound position errors under specified fault hypotheses and risk budgets, but advanced receiver autonomous integrity monitoring (ARAIM) is formulated for snapshot GNSS estimation rather than the correlated posterior of tightly coupled GNSS/IMU factor graph optimization (FGO). This paper formulates ARAIM solution separation and PL computation on the FGO posterior. Because current-epoch pseudorange factors enter only the current-state block of the fixed-lag information matrix, the fault-excluded position covariance, solution separation, and nominal-bias propagation follow exactly from the linearized current-state marginal, which retains correlations with velocity, IMU-bias, and clock states and with historical information. Woodbury updates with selected-output evaluation compute these quantities for fault-free, single-satellite, constellation-wide, and second-order hypotheses without repeated full-state inversion or graph re-optimization. Surrounding monitoring layers screen measurements, handle post-fit outliers, and determine PL validity and availability. In two road experiments, both nominal sequences achieve 100% horizontal and vertical empirical bounding over PL-valid epochs, with a maximum error-to-PL ratio of 0.96. Normalized errors show that the posterior covariance alone does not capture slowly varying ranging biases: a covariance-only bound computed from the same posteriors retains no resolvable margin, and strong temporal correlation limits the empirical evidence to consistency checks rather than validation of the integrity budget. On identical Experiment A posteriors, the proposed method reduces fault-hypothesis evaluation and PL solving time by 65.2% relative to direct full-state inversion with matching integrity decisions, and controlled fault injections demonstrate fail-safe PL invalidation while navigation continues.**

## I. Introduction

TIGHTLY coupled GNSS/IMU factor graph optimization (FGO) represents pseudorange, Doppler, IMU pre-integration, and receiver-clock constraints as factors over a multi-epoch state. The resulting posterior exploits temporal correlations among pose, velocity, inertial-bias, and clock states, while incremental smoothers such as iSAM2 keep the computation real time [1]–[4]. For safety-critical navigation, however, accuracy is not sufficient: a protection level (PL) must bound the position error at a specified integrity risk, and the solution is operationally available only when the PL remains below the horizontal or vertical alert limit (HAL or VAL) [5]. In urban areas, signal blockage, multipath, and non-line-of-sight (NLOS) reception cause frequent and often simultaneous pseudorange errors, so integrity concepts developed for open-sky aviation cannot be transferred directly to ground vehicles [6], [7].

Receiver autonomous integrity monitoring (RAIM) and advanced RAIM (ARAIM) bound GNSS errors through fault hypotheses, solution separation, nominal-bias bounds, and integrity-risk allocation [8], [9], [10], but baseline ARAIM is formulated for a GNSS-only weighted least-squares (WLS) snapshot estimator. Tightly coupled FGO instead couples states across epochs [2], [11], so the effect of a measurement fault depends on the smoothing posterior [12]. Graph-based integrity studies have developed sliding-window fault detection and exclusion (FDE), residual-based fault isolation, fault-dependent error bounds, and robust reweighting [13]–[17], [18]. The problem addressed here is the consistent construction and efficient evaluation of ARAIM solution separation, fault-excluded covariance, nominal-bias propagation, and risk allocation for GNSS pseudorange hypotheses in the correlated GNSS/IMU posterior. The main contributions are:

1) A posterior-domain ARAIM solution-separation and PL formulation for single-satellite, constellation-wide, and second-order current-epoch pseudorange hypotheses in tightly coupled GNSS/IMU FGO. Because these measurements act only on the current state, the fault-excluded covariance, solution separation, and nominal-bias propagation follow exactly from the linearized current-state marginal, and Woodbury updates with selected-output evaluation compute them without full-state inversion or graph re-optimization for each hypothesis.
2) A six-layer integrity-monitoring architecture that combines snapshot FDE, pre-fit screening, post-fit factor removal and state re-estimation, posterior PL computation, cross-system consistency checks, and the final availability decision.

Jiahao Hu, Xiangru Wang, and Weisong Wen are with the Department of Aeronautical and Aviation Engineering, The Hong Kong Polytechnic University, Hong Kong, China.

Qian Meng is with the School of Instrument Science and Engineering, Southeast University, Nanjing 210096, China.

Jiale Wang is with Beihang University, Beijing, China.

Shizhuang Wang is with Shanghai Jiao Tong University, Shanghai, China.

Corresponding author: Weisong Wen.

3) Two road experiments with controlled fault injections that compare the proposed PL with standalone SPP RAIM and a covariance-only bound, quantify bound margins and their statistical resolution under temporally correlated errors, and evaluate fault response, computational cost, nonlinear agreement, and fault-hypothesis coverage.

*Notation:* Scalars are italic; vectors and matrices are bold upright lowercase and uppercase letters, respectively, e.g., $\mathbf{v}_k$, $\mathbf{H}_i$, and $\boldsymbol{\Sigma}_x$. Uppercase italic letters are kept for established scalars such as probabilities, thresholds, and the measurement variance $R_i$, and calligraphic letters denote sets. The superscripts $(0)$ and $(-m)$ mark the all-in-view and fault-excluded solutions, and $Q(\cdot)$ and $\Phi^{-1}(\cdot)$ are the standard-normal upper-tail probability and inverse cumulative distribution function.

## II. Related Work

*RAIM and ARAIM:* Residual-based RAIM originated with Parkinson and Axelrad and with Brown [8], [19]. Baseline ARAIM combines fault hypotheses, solution separation, nominal-bias bounds, and integrity-risk allocation for WLS snapshot estimation [9]. Joerger et al. derived residual-based and solution-separation RAIM from common integrity- and continuity-risk definitions and extended both to FDE [10], [20]. Blanch, Walter, and Enge avoided explicit subset solutions through Woodbury-based subset-variance expressions [21], the computational precedent for this work. Later work addresses fault grouping, temporal exposure, and exclusion [22], [23], and urban ARAIM variants use 3D-city-model NLOS detection or multipath/NLOS fault modes formed by vector-angle grouping [24], [25].

*Integrity of GNSS/INS filters:* In integrated navigation, faults propagate through state prediction and measurement updates. Joerger and Pervan bounded the risk of undetected faults with a Kalman-filter detector built from current and past residuals [26], and Tanil et al. computed worst-case failure slopes recursively from the innovations of a single filter [27]. Kalman-filter solution separation has been recast as a pseudo-snapshot WLS problem [28], and Lee et al. derived detectors and PLs for IMU faults propagated through prediction [29]. These statistics rely on recursive prediction and update; in FGO, the fault effect must instead be evaluated through the state correlations of the smoothing posterior.

*Integrity of factor-graph estimators:* FGO computes the multi-epoch posterior with incremental Bayes-tree updates and on-manifold IMU pre-integration [30] and has been applied to tightly coupled raw-GNSS/inertial urban navigation [11]. Hafez et al. developed solution-separation integrity monitoring for landmark-based fixed-lag localization [12], and Bhamidipati and Gao computed Graph-SLAM PLs through worst-case failure slopes [18]. For GNSS, proposed methods include sliding-window FDE [13]; real-time and cumulative chi-squared tests with GNSS/IMU fault isolation and PL evaluation [14]; single-fault PLs under integrity-constrained weighting [16]; sensor-level fault enumeration with characteristic-slope bounds [15]; and dynamic covariance scaling for robust estimation, which differs from covariance inflation for PL overbounding [17]. A remaining need is an ARAIM formulation that jointly defines solution separation, fault-excluded covariance, nominal-bias propagation, and risk allocation for satellite- and constellation-level pseudorange hypotheses in the correlated GNSS/IMU posterior and evaluates them without repeated full-state inversion.

## III. System Architecture and Factor Graph

Fig. 1 shows measurement processing, incremental fixed-lag smoothing, and six integrity-monitoring layers. Pseudorange, Doppler, and IMU factors constrain the fused state; the single-point-positioning (SPP) solution provides auxiliary consistency information but is not independent because it shares the GNSS observations. Layers 0–1 select current pseudoranges for FGO and fault hypotheses, Layer 2 removes localized post-fit outliers and re-estimates the state, Layer 3 computes PLs from the resulting posterior, Layer 4 supplies auxiliary warnings, and Layer 5 determines operational availability.

*States and factors:* The state at GNSS epoch $k$ is $\mathcal{X}_k = \{\mathbf{T}_k, \mathbf{v}_k, \mathbf{b}_k, \mathbf{c}_k, \dot{\mathbf{c}}_k\}$, where $\mathbf{T}_k = (\mathbf{R}_k, \mathbf{p}_k) \in \mathrm{SE}(3)$ is the IMU/body pose, with $\mathbf{R}_k$ mapping body-frame vectors to the local East-North-Up (ENU) frame and $\mathbf{p}_k$ the ENU position; $\mathbf{v}_k$ is the velocity; $\mathbf{b}_k$ stacks the accelerometer and gyroscope biases $\mathbf{b}_k^a$ and $\mathbf{b}_k^g$; and $\mathbf{c}_k$ and $\dot{\mathbf{c}}_k$ contain the per-constellation receiver clock biases and drifts in range-equivalent units. The window state set is $\mathcal{X}$, and all measurements and priors are collected in $\mathcal{Z}$. The IMU factor uses on-manifold pre-integration of the inertial samples between GNSS epochs with first-order bias correction [30]; its residual $\mathbf{r}_{\mathrm{IMU},k}$ compares the relative rotation, velocity, and position of consecutive states with the pre-integrated increments and models the biases as random walks, with covariance $\boldsymbol{\Sigma}_{\mathrm{IMU},k}$. The clock factor $\mathbf{r}_{\mathrm{clock},k} = [(\mathbf{c}_k - \mathbf{c}_{k-1} - \dot{\mathbf{c}}_{k-1}\Delta t)^T, (\dot{\mathbf{c}}_k - \dot{\mathbf{c}}_{k-1})^T]^T$ applies a constant-drift model with a drift random walk over the epoch interval $\Delta t$. For satellite $i$ of constellation $s(i)$, the pseudorange and Doppler residuals, defined as measured minus predicted values, are

$$\begin{aligned} r_{\rho,i}(k) &= \rho_{i,k}^{\mathrm{meas}} - \left[\left\|\mathbf{p}_{i,k}^{\mathrm{sv}} - \mathbf{p}_k^{\mathrm{ant}}\right\| + c_{k,s(i)} + d_{i,k}\right], \\ r_{D,i}(k) &= \dot{\rho}_{i,k}^{\mathrm{meas}} - [\boldsymbol{\ell}_{i,k}^T(\mathbf{v}_{i,k}^{\mathrm{sv}} - \mathbf{v}_k) \\ &\quad + \dot{c}_{k,s(i)} - c\,\delta\dot{t}_{i,k}^{\mathrm{sv}}], \end{aligned} \tag{1}$$

where $\mathbf{p}_k^{\mathrm{ant}} = \mathbf{p}_k + \mathbf{R}_k \mathbf{l}^{\mathrm{ant}}$ is the antenna position with body-frame lever arm $\mathbf{l}^{\mathrm{ant}}$; $\mathbf{p}_{i,k}^{\mathrm{sv}}$ and $\mathbf{v}_{i,k}^{\mathrm{sv}}$ are the Sagnac-corrected satellite position and the satellite velocity; $d_{i,k} = -c(\delta t_{i,k}^{\mathrm{sv}} - \mathrm{TGD}_i) + d_{i,k}^{\mathrm{trop}} + d_{i,k}^{\mathrm{ion}}$ collects the broadcast satellite clock and group-delay corrections and the modeled tropospheric and ionospheric delays, with $c$ the speed of light; $\boldsymbol{\ell}_{i,k}$ is the receiver-to-satellite unit vector; $\delta\dot{t}_{i,k}^{\mathrm{sv}}$ is the satellite clock drift; and $\dot{\rho}_{i,k}^{\mathrm{meas}} = -\lambda_i f_{D,i,k}$ is the range rate from the Doppler shift $f_{D,i,k}$ at wavelength $\lambda_i$. The lever-arm-induced antenna velocity is neglected.

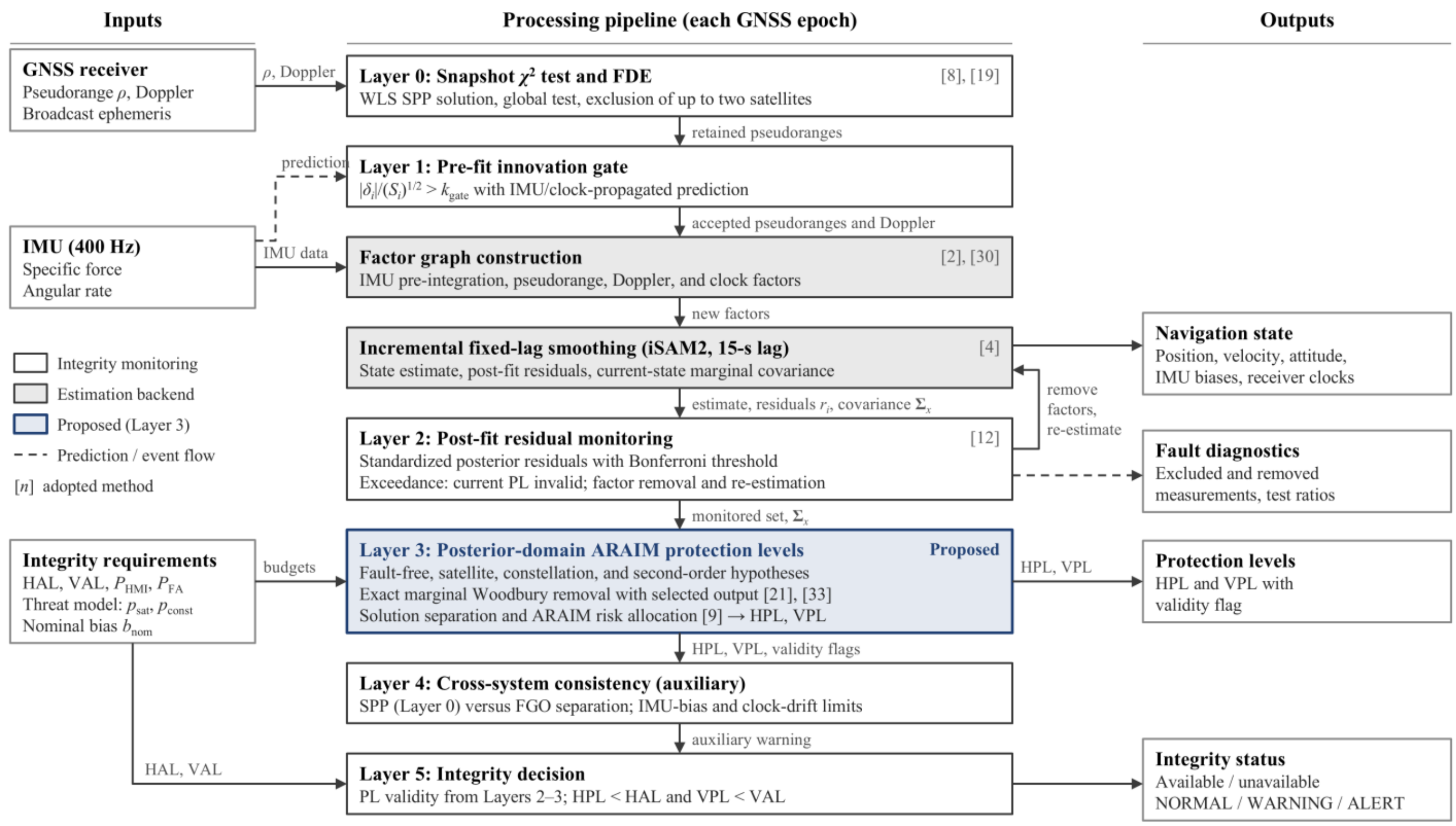

Fig. 1. Architecture of the proposed tightly coupled GNSS/IMU FGO system with layered integrity monitoring. Layers 0–1 screen current-epoch pseudoranges before graph construction; a Layer-2 exceedance invalidates the current PL and triggers re-estimation after factor removal; Layer 3 computes HPL/VPL from the resulting posterior; Layer 4 issues auxiliary warnings only; and Layer 5 combines PL validity and alert-limit tests while navigation continues. Arrow labels give the data passed between blocks; bracketed numbers identify modules adopted from the cited methods, and the shaded Layer-3 block is the proposed posterior-domain ARAIM module.

*Window estimation:* The estimate $\hat{\mathcal{X}}$ maximizes $p(\mathcal{X} \mid \mathcal{Z}) \propto \exp(-1/2\, J_k(\mathcal{X}))$, i.e., minimizes the active-window objective

$$
\begin{aligned}
J_k(\mathcal{X}) = &\sum_j \left\|\mathbf{r}_{\mathrm{IMU},j}\right\|^2_{\boldsymbol{\Sigma}_{\mathrm{IMU},j}} + \sum_j \left\|\mathbf{r}_{\mathrm{clock},j}\right\|^2_{\boldsymbol{\Sigma}_{\mathrm{clk}}} \\
&+ \sum_{j,i} \left\|r_{\rho,i}(j)\right\|^2_{\sigma^2_{\rho,i,j}} + \sum_{j,i} \left\|r_{D,i}(j)\right\|^2_{\sigma^2_{D,i,j}} \\
&+ J_{\mathrm{prior},k}(\mathcal{X}),
\end{aligned} \tag{2}
$$

where $j$ indexes window epochs and $i$ satellites, $\|\mathbf{r}\|^2_{\boldsymbol{\Sigma}} = \mathbf{r}^T \boldsymbol{\Sigma}^{-1} \mathbf{r}$, $\boldsymbol{\Sigma}_{\mathrm{clk}}$ is the clock-factor covariance, and $J_{\mathrm{prior},k}$ collects the initialization constraints and the marginalization prior, which carries historical information and cross-state correlations into the posterior. Both GNSS variances follow an elevation- and carrier-to-noise-density-dependent model,

$$
\sigma^2_{\star,i,j} = \frac{a_\star^2\left[1 + 10^{(C_{\mathrm{ref}} - C_{i,j})/10}\right]}{\sin^2 \varepsilon_{i,j}}, \qquad \star \in \{\rho, D\}, \tag{3}
$$

where $\varepsilon_{i,j}$ is the satellite elevation, $C_{i,j}$ is the carrier-to-noise density in dB-Hz, $C_{\mathrm{ref}} = 45$ dB-Hz, and $a_\rho$ and $a_D$ are base standard deviations (Table II). An incremental fixed-lag smoother with a 15-s lag and Bayes-tree updates [4] marginalizes states leaving the window into this prior. Integrity monitoring uses the current estimate $\hat{\mathcal{X}}_k$ and its full marginal covariance $\boldsymbol{\Sigma}_x$ (Section IV-B1).

## IV. Integrity Monitoring Framework

### A. Layers 0–2: Measurement Screening and Outlier Removal

*Layer 0:* Classical snapshot RAIM [8], [19] screens current-epoch pseudoranges before graph construction. A temporary WLS solution with post-fit residual vector $\mathbf{r}_{\mathrm{SPP}}$ and diagonal inverse measurement covariance $\mathbf{W}$ is tested by

$$
T_{\chi^2} = \mathbf{r}_{\mathrm{SPP}}^T \mathbf{W} \mathbf{r}_{\mathrm{SPP}} > \gamma_0 = F^{-1}_{\chi^2_{\nu_0}}\left(1 - P_{\mathrm{FA}}^{(0)}\right), \tag{4}
$$

where $F^{-1}_{\chi^2_{\nu_0}}$ is the inverse chi-squared distribution function with $\nu_0 = n_0 - d_0$ degrees of freedom for $n_0$ observations and $d_0$ snapshot states, and $P_{\mathrm{FA}}^{(0)}$ is the false-alarm setting. A failed test excludes the satellite with the largest absolute normalized residual $|r_{\mathrm{SPP},i}|/\sqrt{\mathrm{Var}(r_{\mathrm{SPP},i})}$, and the test is repeated until it passes, two satellites have been excluded, or no redundancy would remain. Because sequential exclusion can remove the wrong measurement when several signals are simultaneously affected by NLOS reception or multipath [31], Layer 0 serves only as a pre-screen, and the retained measurements are re-examined by Layers 1–3.

*Layer 1:* Each current pseudorange is tested against a prediction formed before graph insertion. IMU propagation and the clock extrapolation $c_{k,s}^{-} = c_{k-1,s} + \Delta t\, \dot{c}_{k-1,s}$ give the predicted state $\hat{\mathbf{x}}_k^{-}$, and the previous joint marginal $\mathbf{\Sigma}_{x,k-1}$ is propagated with the linearized IMU/clock transition $\mathbf{F}_k$ and process covariance $\mathbf{Q}_k$. The innovation and its variance are

$$\begin{aligned} \delta_i &= \rho_i^{\text{meas}} - h_i(\hat{\mathbf{x}}_k^{-}), \\ S_i &= R_i + \mathbf{H}_i^{-}\left(\mathbf{F}_k \mathbf{\Sigma}_{x,k-1} \mathbf{F}_k^T + \mathbf{Q}_k\right)(\mathbf{H}_i^{-})^T, \end{aligned} \tag{5}$$

where $h_i$ is the pseudorange model, $\mathbf{H}_i^{-}$ its Jacobian at the predicted state, and $R_i$ the measurement variance, so that $S_i$ retains pose–clock correlations. The pseudorange is rejected if $|\delta_i|/\sqrt{S_i} > k_{\text{gate}} = 5$, while the Doppler observation of the same satellite is kept. Layer 1 is a pre-fit consistency screen and does not identify the sensor responsible for an inconsistency.

*Layer 2:* After smoothing, each post-fit pseudorange residual $r_i$ is standardized by its posterior variance and tested with a Bonferroni-adjusted threshold over the $n$ tested pseudoranges:

$$\begin{aligned} z_i^{\text{post}} &= \frac{r_i}{\sqrt{R_i - \mathbf{H}_i \mathbf{\Sigma}_x \mathbf{H}_i^T}}, \\ \mathcal{O}_{2,k} &= \left\{ i : \left|z_i^{\text{post}}\right| > \Phi^{-1}\left(1 - \frac{P_{\text{fa}}}{2n}\right) \right\}, \end{aligned} \tag{6}$$

where $\mathbf{H}_i$ is the current-state Jacobian (Section IV-B1) and $P_{\text{fa}}$ the epoch-level false-alarm allocation; $\mathbf{H}_i \mathbf{\Sigma}_x \mathbf{H}_i^T$ is subtracted because the measurement is included in the estimate [12]. If $\mathcal{O}_{2,k} \neq \varnothing$, the identified current-epoch factors are removed from (2), earlier measurements of the same satellites are retained, and the window is re-optimized, either by an incremental downdate or by relinearization that keeps the marginalization prior at its original linearization point. Layer 3 then uses the re-estimated posterior, but the Layer-2 pass indicator $a_{2,k} = \mathbf{1}(\mathcal{O}_{2,k} = \varnothing)$ is zero and the current PL is reported as unavailable; later epochs are evaluated from their own residuals. Layer 2 therefore combines fail-safe PL invalidation with targeted state re-estimation.

### *B. Layer 3: ARAIM Fusion Protection Levels*

Layer 3 converts the fixed-lag posterior into ARAIM protection levels for the monitored current-epoch pseudorange factors.

*1) From WLS to FGO:* Classical ARAIM uses the covariance and measurement sensitivities of a linearized WLS snapshot estimator; here they are replaced by the posterior of the current full state. Under the right (body-frame) pose perturbation $\mathbf{T}_k' = \mathbf{T}_k \operatorname{Exp}([\delta\boldsymbol{\theta}_k^T, \delta\boldsymbol{\rho}_k^T]^T)$, the current state perturbation is

$$\delta\mathbf{x}_k = [\delta\boldsymbol{\theta}_k^T \quad \delta\boldsymbol{\rho}_k^T \quad \delta\mathbf{v}_k^T \quad \delta\mathbf{b}_k^T \quad \delta\mathbf{c}_k^T \quad \delta\dot{\mathbf{c}}_k^T]^T, \tag{7}$$

with rotational, translational, and velocity perturbations $\delta\boldsymbol{\theta}_k, \delta\boldsymbol{\rho}_k, \delta\mathbf{v}_k \in \mathbb{R}^3$, IMU-bias perturbation $\delta\mathbf{b}_k \in \mathbb{R}^6$, and clock perturbations $\delta\mathbf{c}_k, \delta\dot{\mathbf{c}}_k \in \mathbb{R}^4$. Its covariance $\mathbf{\Sigma}_x = \operatorname{Cov}(\delta\mathbf{x}_k) \in \mathbb{R}^{23\times 23}$ is the current-state marginal of the fixed-lag posterior and therefore incorporates historical factors, marginalization priors, and cross-state correlations. The protected point is the IMU/body origin, whose first-order ENU displacement is $\mathbf{R}_k \delta\boldsymbol{\rho}_k$; the position selector and the all-in-view position covariance are

$$\mathbf{E}_p = [\mathbf{0}_{3\times3} \quad \mathbf{R}_k \quad \mathbf{0}_{3\times17}], \qquad \mathbf{\Sigma}_p^{(0)} = \mathbf{E}_p \mathbf{\Sigma}_x \mathbf{E}_p^T, \tag{8}$$

where $\mathbf{0}_{r\times c}$ is an $r \times c$ zero matrix. A body-fixed point $\mathbf{l}$ would add $-\mathbf{R}_k[\mathbf{l}]_\times$ in the first block; the antenna lever arm enters only the GNSS factor Jacobians. For each monitored pseudorange factor $i$ at the current epoch, with unwhitened post-fit residual $r_i$ and variance $R_i = \sigma_{\rho,i,k}^2$, the information contribution is $\mathbf{H}_i^T R_i^{-1} \mathbf{H}_i$ with full-state Jacobian

$$\mathbf{H}_i = \frac{\partial r_i}{\partial \delta\mathbf{x}_k} = [\mathbf{H}_{T_i} \quad \mathbf{0}_{1\times3} \quad \mathbf{0}_{1\times6} \quad \mathbf{H}_{c_i} \quad \mathbf{0}_{1\times4}], \tag{9}$$

where $\mathbf{H}_{T_i} \in \mathbb{R}^{1\times6}$ is the pose Jacobian and $\mathbf{H}_{c_i} \in \mathbb{R}^{1\times4}$ the clock-bias Jacobian, which is nonzero only in the entry of constellation $s(i)$. Velocity, IMU bias, and clock drift therefore affect Layer 3 only through $\mathbf{\Sigma}_x$. For axis $q \in \{E, N, U\}$, $\mathbf{e}_q \in \mathbb{R}^3$ is the corresponding unit vector, so that $\mathbf{e}_q^T \mathbf{\Sigma}_p \mathbf{e}_q$ is the variance of a position covariance $\mathbf{\Sigma}_p$ along $q$.

*2) ARAIM Fault Hypotheses:* Hypotheses are defined over the current-epoch pseudoranges retained after Layers 0–2. A fault of each visible satellite $i$ and a common fault of each visible constellation $s$ are atomic events, with removed-measurement sets $\mathcal{B}_i^{\text{sat}} = \{i\}$ and $\mathcal{B}_s^{\text{const}} = \mathcal{J}_s$, where $\mathcal{J}_s$ contains the monitored satellites of constellation $s$; the latter captures constellation-level signal, clock, ephemeris, or correction anomalies. All first- and second-order event combinations are monitored. A second-order hypothesis of events $a$ and $b$ removes $\mathcal{B}_a \cup \mathcal{B}_b$, which covers two-satellite, satellite-plus-constellation, and two-constellation faults, and combinations that remove the same set are merged. The fault-free hypothesis $F_0$ removes no measurement, and a generic non-nominal hypothesis $F_m$ removes $\mathcal{B}_m = \{i_1, \dots, i_{n_m}\}$ with $n_m = |\mathcal{B}_m|$.

With atomic prior probabilities $p_a^{\text{atom}}$ and independent occurrences within an epoch, the probability that exactly the events indexed by a subset $\mathcal{U}$ of the visible catalog occur is

$$p(\mathcal{U}) = \prod_{a\in\mathcal{U}} p_a^{\text{atom}} \prod_{b\notin\mathcal{U}} (1 - p_b^{\text{atom}}). \tag{10}$$

The prior $p_m = \Pr(F_m)$ is the sum of $p(\mathcal{U})$ over all first- and second-order subsets whose removed sets have union $\mathcal{B}_m$, and

$$p_0 = \prod_a (1 - p_a^{\text{atom}}), \qquad P_{\text{not monitored}} = 1 - p_0 - \sum_m p_m, \tag{11}$$

so that $P_{\text{not monitored}}$ contains all third- and higher-order combinations. The catalog is rebuilt from the visible satellites and constellations at every epoch. For each mode, the stacked Jacobian, measurement covariance, and residuals are

$$\begin{aligned} \mathbf{H}_m &= \left[\mathbf{H}_{i_1}^T \quad \cdots \quad \mathbf{H}_{i_{n_m}}^T\right]^T, \quad \mathbf{R}_m = \operatorname{diag}(R_{i_1}, \dots, R_{i_{n_m}}), \\ \mathbf{r}_m &= [r_{i_1} \quad \cdots \quad r_{i_{n_m}}]^T. \end{aligned} \tag{12}$$

*3) Solution Separation:* Solution separation measures the position change caused by excluding $\mathcal{B}_m$. Define

$$\mathbf{M}_m = \mathbf{R}_m - \mathbf{H}_m \boldsymbol{\Sigma}_x \mathbf{H}_m^T, \qquad \mathbf{A}_m = \mathbf{E}_p \boldsymbol{\Sigma}_x \mathbf{H}_m^T, \tag{13}$$

where $\mathbf{M}_m \in \mathbb{R}^{n_m \times n_m}$ is the posterior-removal middle matrix and $\mathbf{A}_m \in \mathbb{R}^{3 \times n_m}$ maps the removed information into the protected position. When $\mathbf{M}_m$ is positive definite, the first-order separation $\Delta\mathbf{p}_m = \hat{\mathbf{p}}^{(-m)} - \hat{\mathbf{p}}^{(0)}$, the detection statistic, and its variance along axis $q$ are

$$\begin{aligned} \Delta\mathbf{p}_m &= \mathbf{A}_m \mathbf{M}_m^{-1} \mathbf{r}_m, \qquad D_{m,q} = |\mathbf{e}_q^T \Delta\mathbf{p}_m|, \\ \sigma_{\mathrm{ss},m,q}^2 &= \mathbf{e}_q^T \mathbf{A}_m \mathbf{M}_m^{-1} \mathbf{A}_m^T \mathbf{e}_q. \end{aligned} \tag{14}$$

Threshold design follows the ARAIM distinction between the continuity model used for detection and the integrity model used for PLs [9], [23]. With the retained factor set and linearization point fixed, and with measurement, process, and prior errors zero-mean Gaussian with the modeled covariances under $F_0$, the signed separation $u_{m,q} = \mathbf{e}_q^T \Delta\mathbf{p}_m$ satisfies $u_{m,q} \mid F_0 \sim \mathcal{N}(0, \sigma_{\mathrm{ss},m,q}^2)$, where $\mathcal{N}$ denotes the normal distribution. The horizontal and vertical per-evaluation false-alert budgets $P_{\mathrm{FA},H}$ and $P_{\mathrm{FA},V}$ are allocated equally over the $N_M$ non-nominal modes, giving the thresholds

$$\begin{aligned} P_{\mathrm{FA},m,E} &= P_{\mathrm{FA},m,N} = \frac{P_{\mathrm{FA},H}}{2N_M}, \qquad P_{\mathrm{FA},m,U} = \frac{P_{\mathrm{FA},V}}{N_M}, \\ T_{m,q} &= K_{\mathrm{fa},m,q} \sigma_{\mathrm{ss},m,q}, \qquad K_{\mathrm{fa},m,q} = \Phi^{-1}\left(1 - \frac{P_{\mathrm{FA},m,q}}{2}\right). \end{aligned} \tag{15}$$

Under the continuity model, $\Pr_{\mathrm{cont}}(D_{m,q} > T_{m,q} \mid F_0) = 2Q(K_{\mathrm{fa},m,q}) = P_{\mathrm{FA},m,q}$ for $\sigma_{\mathrm{ss},m,q} > 0$, so by the union bound the channel false-alert probability does not exceed its budget without assuming independence between tests. If any $D_{m,q} > T_{m,q}$, HPL and VPL are unavailable for that epoch while the navigation state remains available; otherwise $T_{m,q}$, not the observed $D_{m,q}$, enters the integrity-risk equation. All hypothesis classes use these equations; only $\mathcal{B}_m$ changes.

*4) Per-Fault Covariance via Rank Updates:* **Direct full-state removal.** With the all-in-view information matrix $\boldsymbol{\Lambda}_x = \boldsymbol{\Sigma}_x^{-1}$ formed once per epoch, the direct-inversion baseline ("dir") computes for each hypothesis

$$\begin{aligned} \boldsymbol{\Lambda}_x^{(-m)} &= \boldsymbol{\Lambda}_x - \mathbf{H}_m^T \mathbf{R}_m^{-1} \mathbf{H}_m, \\ \boldsymbol{\Sigma}_{p,\mathrm{dir}}^{(-m)} &= \mathbf{E}_p \left(\boldsymbol{\Lambda}_x^{(-m)}\right)^{-1} \mathbf{E}_p^T. \end{aligned} \tag{16}$$

**Woodbury formulation.** Using the identity already employed in GNSS WLS ARAIM [21], the fault-excluded covariance follows from the FGO posterior as

$$\boldsymbol{\Sigma}_x^{(-m)} = \boldsymbol{\Sigma}_x + \boldsymbol{\Sigma}_x \mathbf{H}_m^T \mathbf{M}_m^{-1} \mathbf{H}_m \boldsymbol{\Sigma}_x, \tag{17}$$

where $\boldsymbol{\Sigma}_x$ is the correlated current-state marginal rather than a covariance derived from snapshot GNSS geometry.

**Exactness of marginal-domain removal.** Partition the linearized information matrix of the fixed-lag problem as

$$\boldsymbol{\Omega} = \begin{bmatrix} \boldsymbol{\Omega}_{oo} & \boldsymbol{\Omega}_{ox} \\ \boldsymbol{\Omega}_{xo} & \boldsymbol{\Omega}_{xx} \end{bmatrix}, \tag{18}$$

where $x$ denotes the current state and $o$ the other window states, including the variables constrained by marginalization priors. The current-state marginal information is the Schur complement $\boldsymbol{\Lambda}_x = \boldsymbol{\Omega}_{xx} - \boldsymbol{\Omega}_{xo} \boldsymbol{\Omega}_{oo}^{-1} \boldsymbol{\Omega}_{ox}$. The pseudorange factors in $\mathcal{B}_m$ depend only on the current pose and clock bias, so their information $\mathbf{H}_m^T \mathbf{R}_m^{-1} \mathbf{H}_m$ and their information-vector contribution appear only in the $\boldsymbol{\Omega}_{xx}$ block and the corresponding subvector. Removing them leaves $\boldsymbol{\Omega}_{oo}$ and $\boldsymbol{\Omega}_{ox}$ unchanged and therefore changes the marginal information and the marginal information vector by exactly the removed terms. Consequently, $\boldsymbol{\Lambda}_x - \mathbf{H}_m^T \mathbf{R}_m^{-1} \mathbf{H}_m$ is the current-state marginal information of the full-window problem without $\mathcal{B}_m$, and the fault-excluded covariance and estimate shift follow from $\boldsymbol{\Sigma}_x$ and the residuals of $\mathcal{B}_m$ alone. The argument does not hold for measurements at earlier epochs, whose information is coupled through $\boldsymbol{\Omega}_{oo}$ and $\boldsymbol{\Omega}_{ox}$ or has been absorbed into marginalization priors (Section V-D5).

**Selected-output evaluation.** Rather than materializing $\boldsymbol{\Sigma}_x^{(-m)}$, the implementation forms only its position covariance and the rows needed for nominal-bias propagation. With $\mathbf{M}_m$ and $\mathbf{A}_m$ defined above, the Sherman–Morrison–Woodbury identity [32], [33] gives

$$\begin{aligned} \boldsymbol{\Sigma}_p^{(-m)} &= \boldsymbol{\Sigma}_p^{(0)} + \mathbf{A}_m \mathbf{M}_m^{-1} \mathbf{A}_m^T, \\ \mathbf{C}_p^{(-m)} &= \mathbf{E}_p \boldsymbol{\Sigma}_x^{(-m)} = \mathbf{E}_p \boldsymbol{\Sigma}_x + \mathbf{A}_m \mathbf{M}_m^{-1} \mathbf{H}_m \boldsymbol{\Sigma}_x. \end{aligned} \tag{19}$$

For a fixed linearization point and Gaussian measurements, these updates are exact when $\boldsymbol{\Lambda}_x^{(-m)}$ remains positive definite, which also makes $\mathbf{M}_m$ positive definite; for nonlinear FGO they are first-order approximations because the reduced graph is not relinearized. Single-satellite modes reduce to a Sherman–Morrison scalar division, and constellation and second-order modes use the same expression with their stacked measurements. With $d_x = 23$, direct inversion requires an $O(d_x^3)$ factorization per hypothesis after forming $\boldsymbol{\Lambda}_x^{(-m)}$, whereas Woodbury removal forms $\boldsymbol{\Sigma}_x \mathbf{H}_m^T$ and $\mathbf{M}_m$ with $O(d_x^2 n_m + d_x n_m^2)$ operations and factorizes the $n_m \times n_m$ matrix $\mathbf{M}_m$ at $O(n_m^3)$ cost. The advantage is therefore largest for small removal sets and decreases as $n_m$ grows toward $d_x$ (Section V-D2). The fault-free and fault-excluded standard deviations along axis $q$ are

$$\sigma_q^{(0)} = \sqrt{\mathbf{e}_q^T \boldsymbol{\Sigma}_p^{(0)} \mathbf{e}_q}, \qquad \sigma_q^{(-m)} = \sqrt{\mathbf{e}_q^T \boldsymbol{\Sigma}_p^{(-m)} \mathbf{e}_q}. \tag{20}$$

*5) Nominal Bias Accumulation:* For integrity-bound design, the pseudorange error is modeled as $\epsilon_i^{\mathrm{int}} = \eta_i + \beta_i$, where $\eta_i \sim \mathcal{N}(0, R_i)$ and $\beta_i$ is an unknown bias with $|\beta_i| \le b_{\mathrm{nom},i}$. Unlike the zero-mean continuity model used for detection, this integrity model allows a nonzero fault-free mean. It is consistent with paired overbounding of nonzero-mean range-error distributions [34], and $\sqrt{R_i}$ and $b_{\mathrm{nom},i}$ should be selected to overbound the empirical error distribution [35]. The fault-free and fault-excluded position sensitivities to measurement $i$ follow from $\mathbf{E}_p \boldsymbol{\Sigma}_x$ and from (19), and the corresponding bias bounds over the monitored set $\mathcal{J}$ are

$$\mathbf{G}_i^{(0)} = \mathbf{E}_p \boldsymbol{\Sigma}_x \mathbf{H}_i^T R_i^{-1}, \qquad \mathbf{G}_i^{(-m)} = \mathbf{C}_p^{(-m)} \mathbf{H}_i^T R_i^{-1},$$
$$b_q^{(0)} = \sum_{i \in \mathcal{J}} | \mathbf{e}_q^T \mathbf{G}_i^{(0)} | \, b_{\text{nom},i},$$
$$b_q^{(-m)} = \sum_{i \in \mathcal{J} \setminus \mathcal{B}_m} | \mathbf{e}_q^T \mathbf{G}_i^{(-m)} | \, b_{\text{nom},i}. \tag{21}$$

These margins enter the PL risk equation, not the detection threshold. They do not guarantee the stated false-alert probability for all admissible $\beta_i$, which can shift the separation mean; no separation-bias margin is added to $T_{m,q}$ in this implementation.

*6) ARAIM Protection-Level Equation:* Following the ARAIM user algorithm [9] with the WLS quantities replaced by the FGO posterior quantities above, the integrity risk for a candidate protection level $\text{PL}_q$ on axis $q$ is bounded by

$$\Psi_q(\text{PL}_q) = 2Q\left(\frac{\text{PL}_q - b_q^{(0)}}{\sigma_q^{(0)}}\right) + \sum_m p_m \, Q\left(\frac{\text{PL}_q - T_{m,q} - b_q^{(-m)}}{\sigma_q^{(-m)}}\right), \tag{22}$$

where the factor 2 accounts for the two-sided fault-free error event. With horizontal and vertical hazardously misleading information (HMI) budgets $P_{\text{HMI},H}$ and $P_{\text{HMI},V}$ and $P_{\text{HMI,total}} = P_{\text{HMI},H} + P_{\text{HMI},V}$, the unmonitored probability is deducted proportionally, and the axis targets are

$$P_{\text{HMI},\kappa}^{\text{eff}} = P_{\text{HMI},\kappa}\left(1 - \frac{P_{\text{not monitored}}}{P_{\text{HMI,total}}}\right), \qquad \kappa \in \{H, V\},$$
$$P_{\text{target},U} = P_{\text{HMI},V}^{\text{eff}}, \qquad P_{\text{target},E} = P_{\text{target},N} = 1/2 \, P_{\text{HMI},H}^{\text{eff}}. \tag{23}$$

The allocation is feasible only when $P_{\text{not monitored}} < P_{\text{HMI,total}}$; otherwise the PL is unavailable. Since $\Psi_q$ decreases monotonically, $\Psi_q(\text{PL}_q) = P_{\text{target},q}$ is solved by bisection over $[0{,}500]$ m with a tolerance of $0.01$ m, returning the risk-feasible upper endpoint; a failed solution or covariance check also makes the PL unavailable. The outputs are

$$\text{VPL} = \text{PL}_U, \qquad \text{HPL} = \sqrt{\text{PL}_E^2 + \text{PL}_N^2}, \tag{24}$$

where the horizontal construction is conservative under the union bound because an error outside the horizontal radius exceeds at least one component bound.

### C. Layers 4–5: Consistency Checks and Integrity Decision

*Layer 4* is an auxiliary estimator-health screen rather than an independent integrity bound, because SPP and FGO share GNSS measurements. It raises a warning when the SPP–FGO position difference, the accelerometer- or gyroscope-bias norm, or the largest constellation clock drift exceeds its threshold $\tau_{\text{SF}}$, $\tau_a$, $\tau_g$, or $\tau_{\dot{c}}$ (Table II); a warning does not alter the Layer-3 risk equation or invalidate an otherwise valid PL.

*Layer 5* separates PL validity from operational availability. With binary indicators $a_{\text{SS}} = 1$ when every test satisfies $D_{m,q} \leq T_{m,q}$, $a_{\text{risk}} = 1$ when $P_{\text{not monitored}} < P_{\text{HMI,total}}$, and $a_{\text{num}} = 1$ when all hypothesis covariances and PL solutions are valid and finite,

$$a_{\text{PL}} = a_{2,k} \wedge a_{\text{SS}} \wedge a_{\text{risk}} \wedge a_{\text{num}},$$
$$a_H = a_{\text{PL}} \wedge (\text{HPL} < \text{HAL}), \qquad a_V = a_{\text{PL}} \wedge (\text{VPL} < \text{VAL}), \tag{25}$$

where $a_{\text{PL}}$ indicates PL validity and $a_H$ and $a_V$ indicate horizontal and vertical operational availability. A valid PL at or above its alert limit remains a numerical bound, but the channel is operationally unavailable; the navigation state remains available in all cases. The experiments report three mutually exclusive states: ALERT for an unavailable PL or an alert-limit exceedance, WARNING for a Layer-4 inconsistency not superseded by an alert, and NORMAL otherwise.

## V. Experimental Results and Discussion

### A. Experiment Conditions and Evaluation Protocol

Two Hong Kong road experiments were evaluated. Experiment A is the primary long-distance sequence for nominal and controlled-fault tests; Experiment B uses UrbanNav-HK-Deep-Urban-1 for supporting evaluation in a dense urban canyon [36]. Figs. 2 and 3 show the equipment and reference trajectories, respectively; Table I summarizes the acquisition settings. All computation was performed on an Intel Core Ultra 7 265KF desktop with 32 GB RAM.

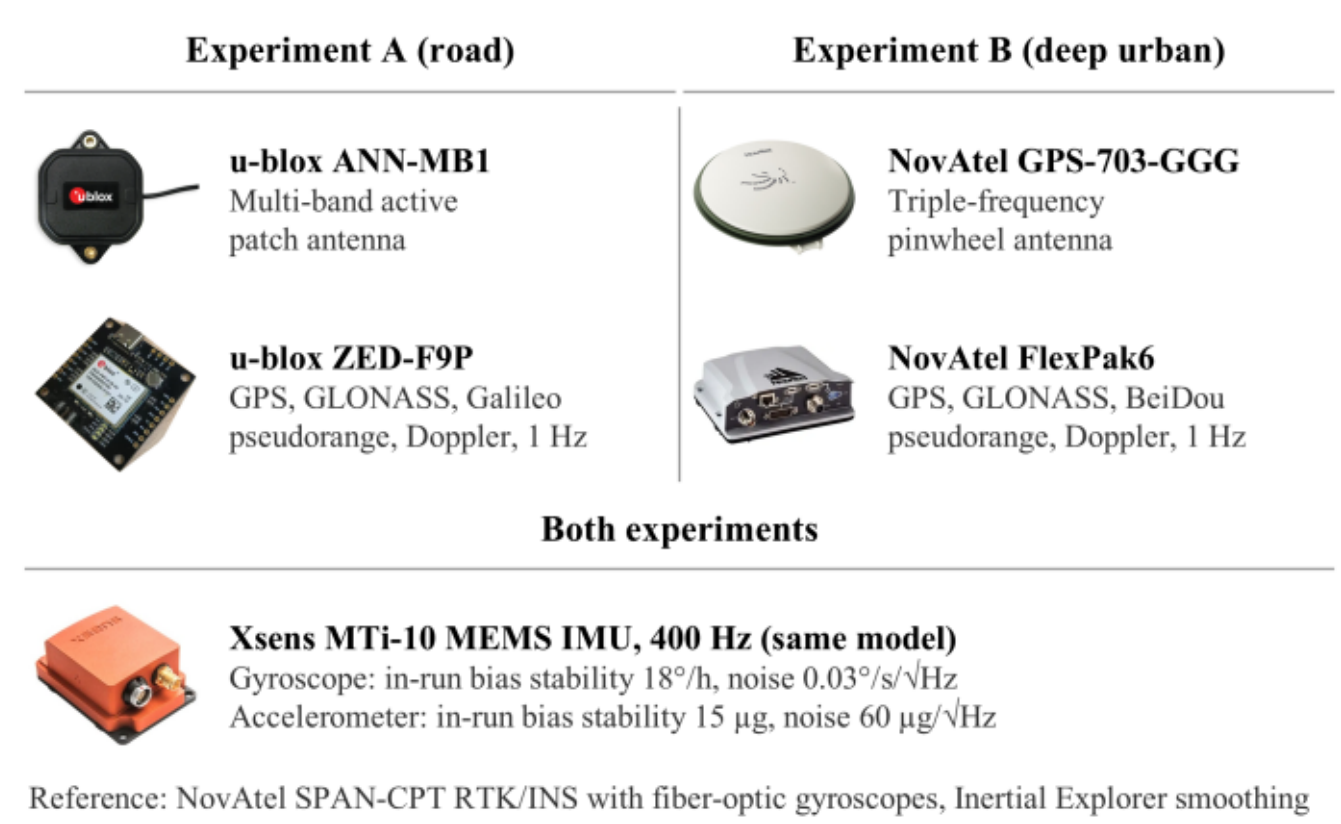

Fig. 2. Sensor hardware and key specifications. The GNSS antenna and receiver differ between Experiments A and B, whereas the same IMU model is used in both. GNSS entries list the constellations and observables used in this study; IMU values are datasheet in-run bias stability and noise density [37].

Reference trajectories were generated from NovAtel SPAN-CPT GNSS/INS measurements with Inertial Explorer (IE) postprocessing [36]; Experiment A uses a 100-Hz smoothed solution. Reference positions are linearly interpolated in Earth-centered Earth-fixed coordinates at estimator timestamps, and errors are resolved in ENU at the reference location without trajectory alignment, using ellipsoidal heights and a zero reference-to-body lever arm; Experiment B retains a fixed $-0.14$ m reference-height adjustment. IE quality numbers 1, 2, and 3 correspond to approximate 3-D accuracies of 0–0.15, 0.05–0.40, and 0.20–1.00 m [38]; at the PL-valid epochs of Experiment A, they occur for 45.5%, 52.9%, and 1.6% of epochs, with no worse quality. The UrbanNav reference uses the same receiver and software, but its accuracy is not guaranteed in deep-urban segments without fixed RTK [36]. Reference uncertainty is thus small relative to meter-level errors and PLs, but not relative to decimeter-level differences between an error and a bound.

TABLE I
EXPERIMENTAL SETUP

| Item | Experiment A | Experiment B |
|---|---|---|
| Role / environment | Primary / road | Supporting / deep urban |
| GNSS receiver | u-blox ZED-F9P | NovAtel FlexPak6 |
| GNSS antenna | u-blox ANN-MB1 | NovAtel GPS-703-GGG |
| Constellations | GPS/GLO/GAL | GPS/GLO/BDS |
| Approx. distance | 21.82 km | 4.51 km |
| Location / platform | Hong Kong, China / test vehicle | |
| GNSS / IMU rate | 1 Hz / 400 Hz | |
| IMU model | Xsens MTi-10 (shared model) | |
| Reference trajectory | NovAtel SPAN-CPT + IE solution | |

Experiment B hardware follows the UrbanNav documentation [36] and dataset repository [39].

Fig. 4 characterizes the satellite geometry of both experiments using IGS combined broadcast ephemerides [40] evaluated at the mean reference position. In Experiment A, 18–20 satellites are above the 15° mask (open-sky PDOP 1.15–1.48), and the estimator uses on average 15.6 pseudoranges per epoch (84%), with temporary drops to 8. In Experiment B, 35–36 satellites are above the mask (PDOP 0.91–0.98), but only 11.7 pseudoranges are used on average (33%, range 0–23), consistent with the nearly 70% sky blockage of this sequence [36]; Fig. 3 shows where along each route these drops occur.

All evaluations use quadratic GNSS measurement losses. Table II lists the integrity settings: satellite- and constellation-fault priors follow the Milestone 2 threat model [41], while nominal-bias bounds are fixed study parameters. Risk budgets are applied per PL evaluation. HAL = 15 m lies within the 10–25 m and 10–20 m ranges identified for loose safety-critical and stringent payment-critical road applications [42]; since these requirements are only horizontal, VAL = 30 m is a looser study setting, and Fig. 8(c) reports availability versus alert limit. The standalone SPP RAIM baseline uses a worst-case single-measurement slope with $P_{\mathrm{FA}}^{\mathrm{SPP}} = 10^{-5}$ and $P_{\mathrm{MD}}^{\mathrm{SPP}} = 10^{-3}$; because its risk model differs, comparisons evaluate standalone and fused systems rather than algorithms at matched integrity risk.

Let $e_H$ denote the horizontal position-error magnitude and $e_V$ the absolute vertical error, both in meters. Empirical bound rates are the percentages of PL-valid epochs satisfying HPL $\geq e_H$ and VPL $\geq e_V$; operational availability additionally requires the PL to remain below the alert limit, and PL-unavailable epochs and missing outputs count as unavailable but are excluded from bound rates.

TABLE II
ESTIMATOR AND INTEGRITY PARAMETERS

| Parameter | Value |
|---|---|
| Fixed-lag window / elevation mask | 15 s / 15° |
| Base std. dev. $a_p$ / $a_D$ in (3) | 2 m / 0.2 m/s |
| Layer-0 $P_{\mathrm{FA}}^{(0)}$ / maximum exclusions | $10^{-5}$ / 2 |
| Layer-1 innovation gate $k_{\mathrm{gate}}$ | 5.0 |
| Layer-2 post-fit $P_{\mathrm{fa}}$ | $10^{-5}$ |
| Layer-2 remediation | Factor removal and re-estimation |
| $P_{\mathrm{FA},H}$ / $P_{\mathrm{FA},V}$ | $2\times10^{-6}$ / $2\times10^{-6}$ |
| $P_{\mathrm{HMI,total}}$ | $10^{-7}$ |
| $P_{\mathrm{HMI},H}$ / $P_{\mathrm{HMI},V}$ | $5\times10^{-8}$ / $5\times10^{-8}$ |
| $p_{\mathrm{sat}}$ / $p_{\mathrm{const}}$ | $10^{-5}$ / $10^{-4}$ |
| Maximum monitored fault order | 2 |
| Nominal bias bound $b_{\mathrm{nom},i}$ | 2 m |
| PL bisection interval / tolerance | [0,500] m / 0.01 m |
| Layer-4 $\tau_{\mathrm{SF}}$ / $\tau_a$ | 25 m / 5 m/s$^2$ |
| Layer-4 $\tau_g$ / $\tau_{\dot{c}}$ | 1 rad/s / 1000 m/s |
| HAL / VAL | 15 m / 30 m |

*B. Primary Experiment A: Nominal PL and Fault Injection*

*1) Nominal Protection-Level Evaluation:* Fig. 5(a) and (b) show the horizontal and vertical errors together with the SPP RAIM and fusion PLs for the Experiment A nominal run; Table III summarizes their PL statistics. One Layer-2 exceedance makes the PL unavailable for one epoch without interrupting navigation, so valid PLs are obtained for 1310 of 1311 epochs (99.92%), with no Layer-3 alert.

All 1310 PL-valid epochs satisfy HPL $\geq e_H$ and VPL $\geq e_V$, giving empirical bound rates of 100.0% in both channels (Fig. 6(a) and (b)). The horizontal and vertical root-mean-square errors (RMSEs) over all 1311 navigation outputs are 7.856 m and 6.568 m, with 95th-percentile errors of 9.803 m and 12.915 m, respectively; the maximum errors at PL-valid epochs are 11.727 m and 18.313 m. On the same epochs, standalone SPP gives RMSEs of 7.966 m and 9.447 m, so the fusion mainly improves the vertical channel in this open road sequence. However, 160 valid HPLs reach or exceed HAL and 16 valid VPLs reach or exceed VAL. Across all 1311 evaluated epochs, operational availability is 87.72% in the horizontal channel and 98.70% in the vertical channel.

To isolate the contribution of the nominal-bias and fault-hypothesis terms, Table III also lists a covariance-only bound. It retains only the fault-free Gaussian term of $\Psi_q$, sets $b_q^{(0)} = 0$, and assigns the full channel budget to that term, so that $\mathrm{PL}_q = K_q\sigma_q^{(0)}$ with $K_E = K_N = Q^{-1}(P_{\mathrm{HMI},H}/4) = 5.57$ and $K_U = Q^{-1}(P_{\mathrm{HMI},V}/2) = 5.45$. On the same posteriors, this bound is exceeded once horizontally: at $t = 570$ s, $e_H = 9.996$ m against a covariance-only HPL of 9.974 m and a proposed HPL of 10.434 m. The 0.02-m exceedance is below the reference uncertainty there (quality 2, 0.05–0.40 m) and is not by itself a bounding failure, but the covariance-only bound has no resolvable margin, whereas the proposed

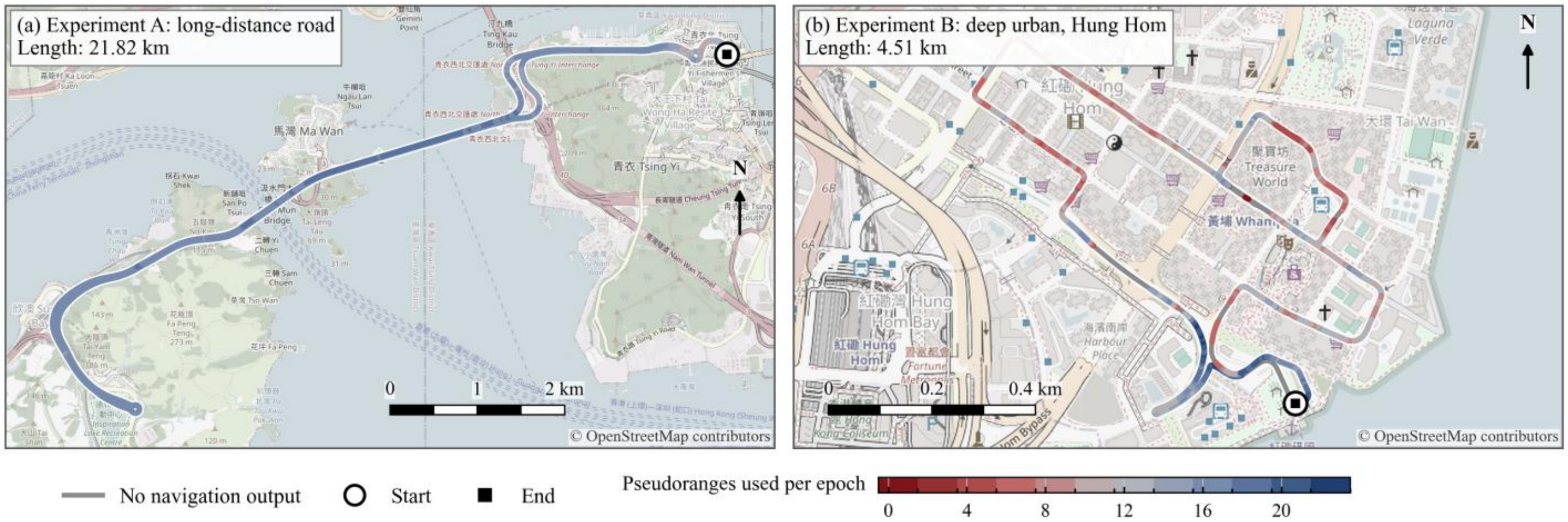

Fig. 3. Reference trajectories on OpenStreetMap: (a) Experiment A long-distance road sequence in Hong Kong, China; (b) Experiment B UrbanNav deep-urban sequence in Hung Hom, Hong Kong, China. In both experiments, the start and end points nearly coincide. Track color gives the number of pseudoranges used by the estimator at each navigation epoch; gray segments have no navigation output. Map data: © OpenStreetMap contributors.

HPL retains 0.44 m; Section V-D1 traces this to time-correlated ranging biases. In the fault-injection run, the covariance-only bound is likewise exceeded at one background epoch and at none within the injection windows. Including the nominal-bias and fault-hypothesis terms raises the mean HPL and VPL by 5.5% and 5.3% and lowers horizontal operational availability from 92.98% to 87.72%; this is the availability cost of protecting against the monitored faults and bounded nominal biases, which the covariance-only bound does not cover.

*2) Controlled Pseudorange-Fault Response and Remediation:* The fault-injection tests evaluate measurement screening, solution-separation response, and PL availability under controlled pseudorange biases (Fig. 7 and Table IV). Biases were injected in a second Experiment A run at 5 × playback, with the nominal run as the no-injection control under otherwise identical estimator, threat-model, and integrity settings. Fault windows and response delays are measured using the original GNSS timestamps.

The four GPS stress cases comprise a 60-m GPS11 step, a GPS11 ramp of 2 m/s capped at 80 m, a differential GPS step with satellite-dependent amplitudes from −80 to +80 m, and an 80-m GPS11/GPS20 step. A separate second-order test applies +30 m to GPS11 and −30 m to GPS20 for one GNSS epoch. The target satellites are continuously present during the corresponding control windows.

The no-injection control has one epoch with unavailable PL and no Layer-3 solution-separation alert. In the 60-m single-GPS and 80-m dual-GPS steps, upstream screening prevents evaluation of the corresponding Layer-3 fault hypotheses; all 30 outputs in each window nevertheless retain valid PLs based on the remaining measurements. The ramp's matching hypothesis reaches $D/T = 0.726$ without crossing its threshold, and one epoch is invalidated by Layer 2. The differential GPS hypothesis reaches $D/T = 1.217$ and produces 12 Layer-3 exceedance epochs, with a first response after six seconds; one additional Layer-2 event gives 13 unavailable PLs in that window. No valid-PL bound violation occurs in any of the four windows.

Across the complete campaign, 1296 of 1311 navigation epochs have valid PLs, with empirical bound rates of 100.0% in both channels. The 15 unavailable outputs comprise three Layer-2 events and 12 Layer-3 events, including one background Layer-2 event outside the injection windows. All three remediation attempts succeed, removing three factors. The mean and 95th-percentile re-estimation times are 4.426 ms and 6.082 ms, respectively, while all 1311 navigation outputs remain present.

In the isolated opposite-sign dual-GPS test, both target measurements pass Layers 0–2, allowing evaluation of the corresponding two-satellite hypothesis. Its $D/T = 0.974$ remains below the detection threshold, so the injected epoch retains valid PLs. The HPL and VPL are 14.101 m and 24.826 m, compared with horizontal and vertical errors of 9.573 m and 1.403 m, respectively. No valid-PL bound violation occurs over the complete run.

Under the tested faults, large steps are thus handled by upstream screening, the differential anomaly that survives screening triggers a solution-separation alert, and a fault below the detection threshold remains bounded by the valid PLs; the current PL is invalidated whenever a residual or separation threshold is exceeded, while navigation output continues.

### *C. Supporting Experiment B: Deep-Urban Validation*

Experiment B evaluates nominal PL bounding in a deep urban canyon. Fig. 5(c)–(d) and Fig. 6(c)–(d) show the time histories and Stanford diagrams. Of 1262 navigation outputs, 1241 provide valid PLs (98.34%). All 1241 valid HPLs and VPLs bound the corresponding reference errors, giving empirical bound rates of 100.0% in both channels. The other 21 outputs lack a valid PL: 20 have no usable current-epoch pseudorange observations and one fails the solution-separation test.

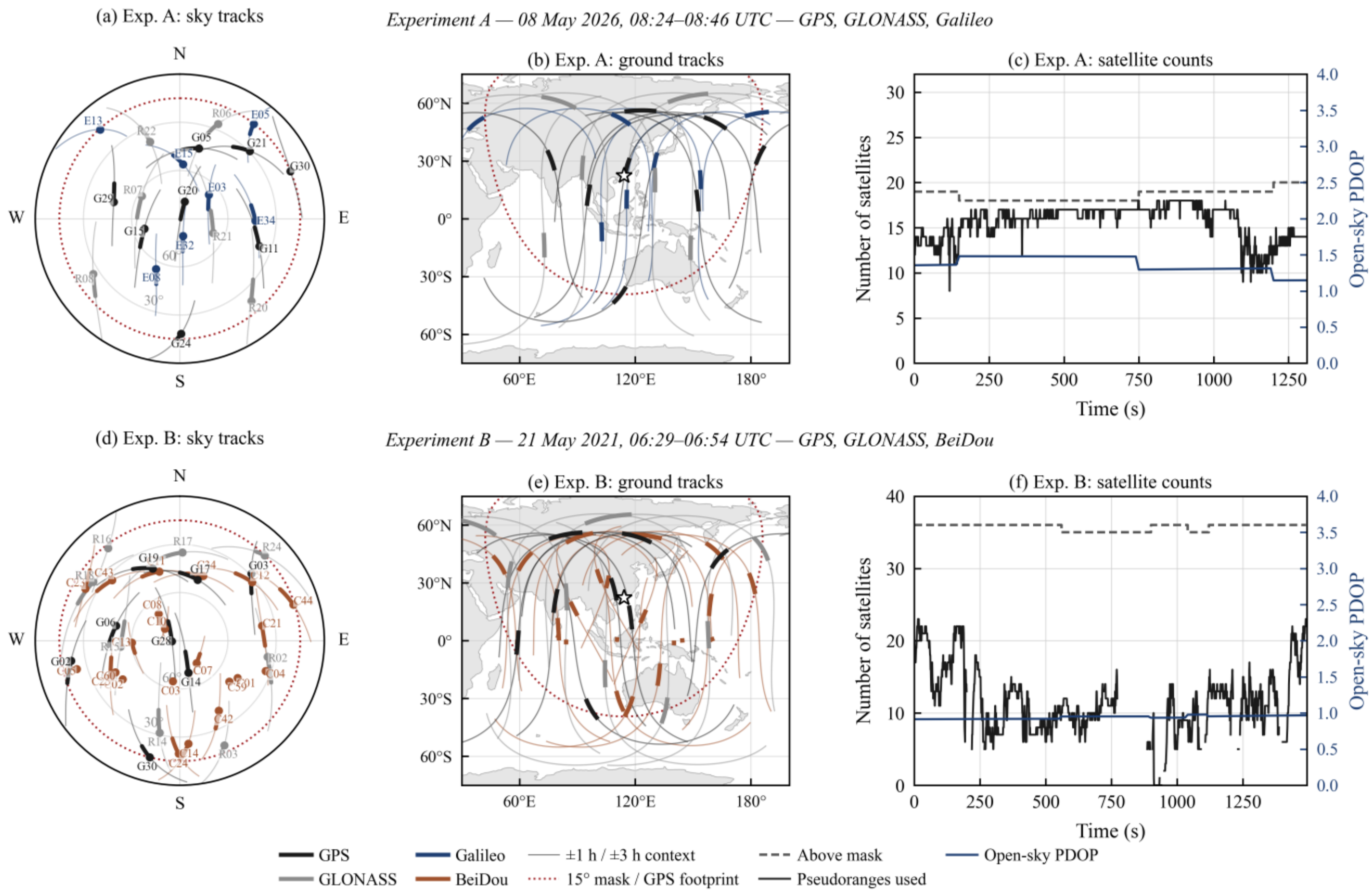

Fig. 4. Satellite geometry from IGS broadcast ephemerides at the mean reference position of each experiment. (a), (d) Sky tracks above the 15° mask during the experiment (thick) and within ±1 h (thin). (b), (e) Ground tracks within ±3 h, experiment interval highlighted; the dotted circle is the 15° GPS visibility footprint around the test area (star). (c), (f) Satellites above the mask (dashed), pseudoranges used after screening (solid), and open-sky PDOP.

Table III lists the PL statistics. Maximum horizontal and vertical errors at PL-valid epochs are 56.497 and 25.265 m; RMSEs over all 1262 outputs are 4.990 and 7.180 m, compared with 15.103 and 20.567 m for standalone SPP on the same epochs. The covariance-only bound defined in Section V-B1 also bounds every valid epoch in this sequence; the nominal-bias and fault-hypothesis terms raise the mean HPL and VPL by 11.7% and 4.0%. However, 930 valid HPLs reach or exceed HAL and 146 valid VPLs reach or exceed VAL, so these outputs are operationally unavailable in the corresponding channel. The full evaluation period contains 1493 epochs at 1 Hz, including 231 without a navigation solution; navigation solutions and valid PLs are available in 84.53% and 83.12% of epochs, and operational availability is 20.83% horizontally and 73.34% vertically.

### D. Cross-Experiment Analysis

Table V summarizes the layer-wise outcomes of the nominal runs. Screening responds differently in the two sequences: Layer 0 dominates in Experiment A and Layer 1 in B. Layer-1 rejections are prediction inconsistencies rather than labeled faults, and Layer-5 ALERT counts combine PL-unavailable events with alert-limit exceedances rather than confirmed sensor faults.

*1) Bound Margins and Statistical Resolution:* Fig. 8(a) shows the exceedance probabilities of the normalized errors $e_H/\sigma_H^{(0)}$ and $e_V/\sigma_U^{(0)}$ at PL-valid epochs, where $\sigma_H^{(0)} = ((\sigma_E^{(0)})^2 + (\sigma_N^{(0)})^2)^{1/2}$; the Gaussian reference $2Q(x)$ is exact vertically and conservative horizontally. In Experiment A, 79.2% of the horizontal normalized errors exceed 3 (95th percentile 4.68), against 0.27% for the reference. The mean East, North, and Up errors are 2.65, 7.21, and 5.26 m (standard deviations 0.84, 1.42, and 3.94 m), and standalone SPP shows almost the same means of 2.60, 6.99, and 5.76 m. This common offset is consistent with slowly varying ranging errors such as residual ionospheric delay after broadcast-model correction, rather than an artifact of the fusion or the reference, and cannot be represented by a posterior covariance that models zero-mean, temporally white pseudorange errors; the horizontal normalized errors therefore approach $K_H = 5.57$. In Experiment B, the horizontal normalized errors are smaller (95th percentile 1.83), but the fused and SPP solutions share a vertical offset of about 5.3 m.

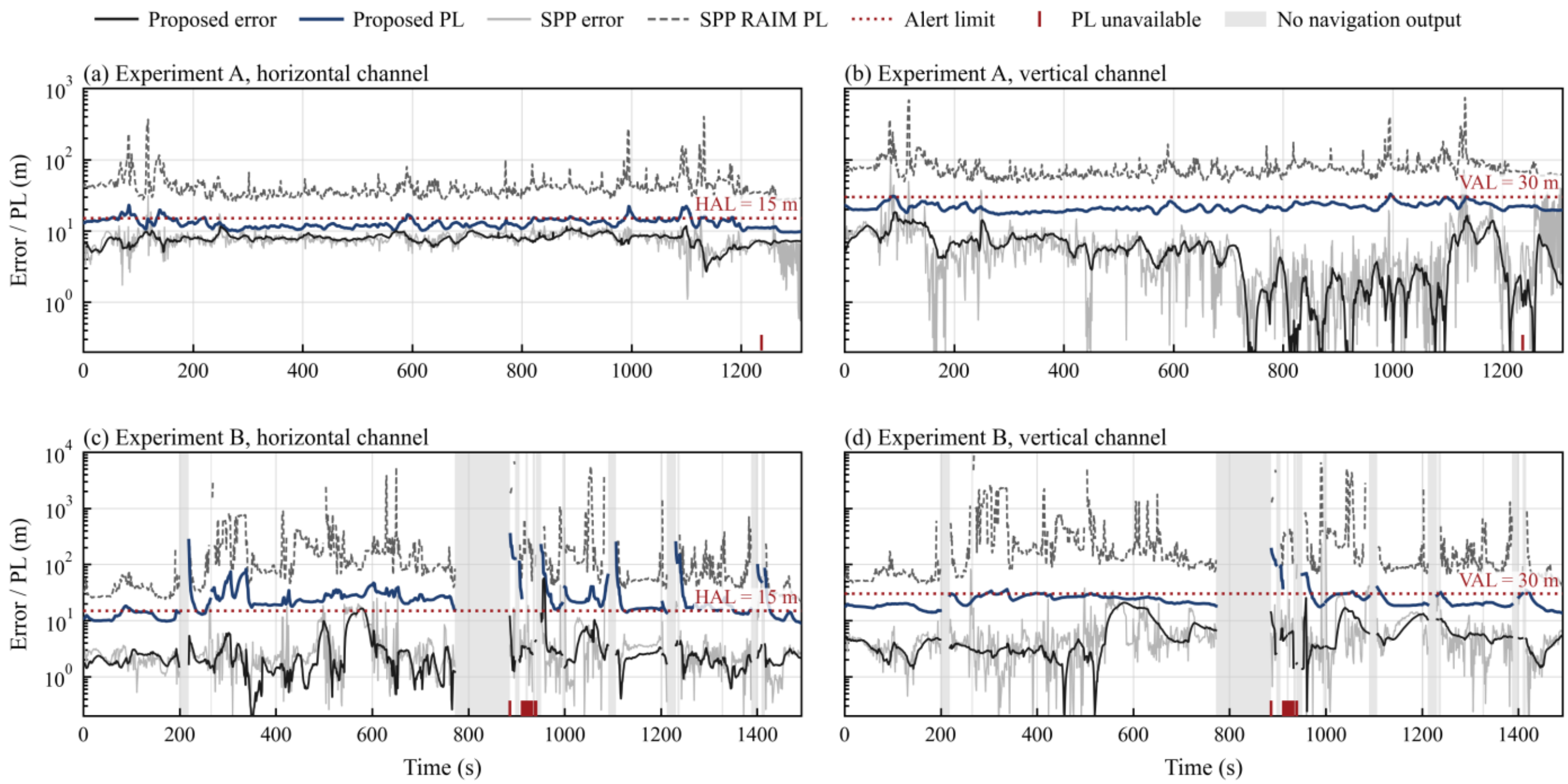

Fig. 5. Position errors and PLs on a logarithmic scale for Experiment A (a, b) and Experiment B (c, d); SPP RAIM and the proposed system use their respective risk models. Proposed PLs are drawn only at PL-valid epochs (1310 in A and 1241 in B); red ticks mark outputs with an unavailable PL, and gray bands mark missing navigation outputs, which are not interpolated. Dotted lines mark HAL = 15 m and VAL = 30 m; values below 0.2 m are drawn at the lower axis limit.

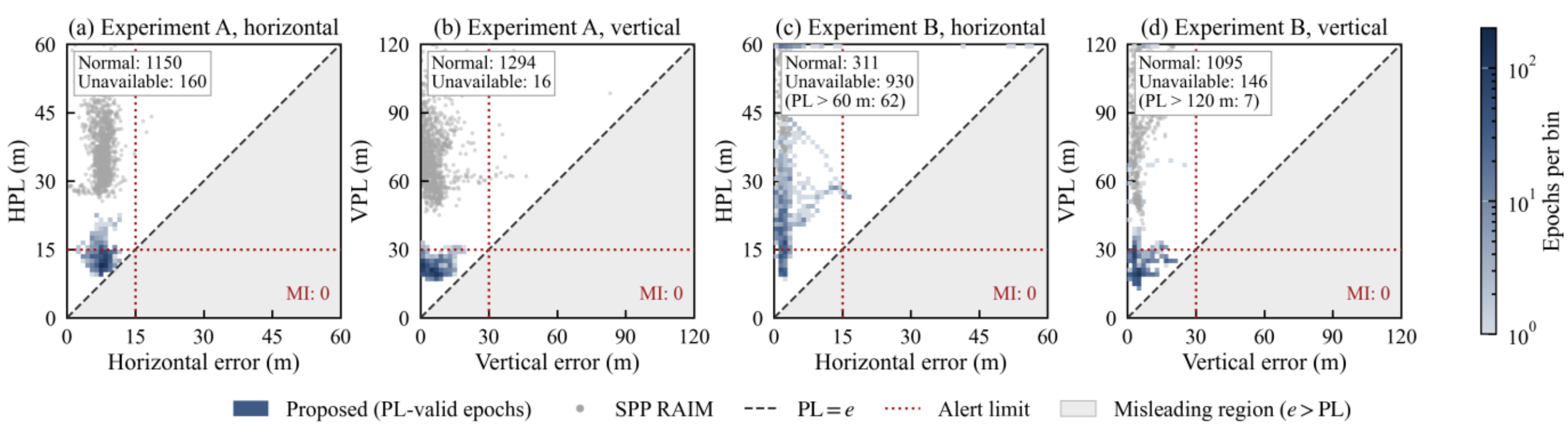

Fig. 6. Stanford diagrams under method-specific risk models: (a,b) Experiment A and (c,d) Experiment B, horizontal and vertical channels. Shaded cells count PL-valid epochs of the proposed system in 1-m (horizontal) and 2-m (vertical) bins; gray dots show SPP RAIM. The misleading-information (MI) region below the diagonal contains no proposed-system epoch. Annotations count normal and operationally unavailable epochs; PLs beyond the axes are placed in the top row.

Fig. 8(b) compares error-to-bound ratios in Experiment A. The maximum ratios are 0.958 for the proposed HPL and 1.002 for the covariance-only HPL, and 0.771 and 0.820 for the corresponding VPLs. The minimum horizontal margin of the proposed PL, 0.44 m, occurs at the same epoch and is comparable to the upper end of the reference uncertainty there. In Experiment B, the maximum ratios of the proposed HPL and VPL are 0.839 and 0.853, with minimum margins of 6.76 and 3.56 m.

The empirical bound rates must also be read in light of the temporal correlation of the errors. The integrated autocorrelation times of the normalized errors are 55 and 139 s for the horizontal and vertical channels of Experiment A and 28 and 52 s for Experiment B. The PL-valid epochs therefore correspond to only about 24, 9, 44, and 24 effectively independent samples [43]. With no exceedance among $n$ independent samples, the per-sample exceedance probability is bounded by approximately $3/n$ at 95% confidence [44], here $0.07-0.32$. The 100% bound rates are thus consistency checks of the error models under the tested conditions and remain many

TABLE III
PROTECTION-LEVEL STATISTICS AND COVARIANCE-ONLY COMPARISON

| | Experiment A | | | | Experiment B | | | |
|---|---|---|---|---|---|---|---|---|
| PL | Mean (m) | 95th pct. (m) | Bound rate | Availability | Mean (m) | 95th pct. (m) | Bound rate | Availability |
| SPP RAIM HPL | 42.896 | 71.314 | 100.00% | 0.00% | 192.860 | 650.297 | 100.00% | 0.00% |
| SPP RAIM VPL | 78.618 | 119.239 | 100.00% | 0.00% | 318.940 | 1531.024 | 100.00% | 0.00% |
| Covariance-only HPL | 12.203 | 15.832 | 99.92% | 92.98% | 23.276 | 50.365 | 100.00% | 24.51% |
| Covariance-only VPL | 20.573 | 25.597 | 100.00% | 99.54% | 23.542 | 31.671 | 100.00% | 76.96% |
| Proposed HPL | 12.876 | 16.711 | 100.00% | 87.72% | 25.997 | 59.907 | 100.00% | 20.83% |
| Proposed VPL | 21.658 | 26.851 | 100.00% | 98.70% | 24.481 | 32.646 | 100.00% | 73.34% |

SPP RAIM and the proposed system use method-specific risk models; the covariance-only bound $\mathrm{PL}_q = K_q\sigma_q^{(0)}$ uses the same posteriors. PL statistics and bound rates use PL-valid epochs (1310 in A and 1241 in B); SPP RAIM uses its finite outputs. Availability uses the 1311 evaluated epochs in A and all 1493 epochs, including missing outputs, in B.

orders of magnitude away from validating the $10^{-7}$ integrity budget; such validation requires overbounding analyses on large independent data sets [35], [45].

TABLE IV
PSEUDORANGE FAULT-INJECTION RESPONSE IN EXPERIMENT A

| Case | Max $D/T$ | Match. $D/T$ | L2 epochs | L3 $D/T > 1$ | PL unavail. | Viol. H/V |
|---|---|---|---|---|---|---|
| Control | 0.820 | – | 1 | 0 | 1 | 0/0 |
| GPS11 step | 0.198 | screened | 0 | 0 | 0 | 0/0 |
| GPS11 ramp | 0.727 | 0.726 | 1 | 0 | 1 | 0/0 |
| Diff. GPS | 1.217 | 1.217 | 1 | 12 | 13 | 0/0 |
| Dual-GPS step | 0.170 | screened | 0 | 0 | 0 | 0/0 |
| Four-case campaign | 1.217 | – | 3 | 12 | 15 | 0/0 |
| Opposite dual-GPS† | 0.974 | 0.974 | 0 | 0 | 0 | 0/0 |

Quadratic loss with Layer-2 factor removal. Match.: largest $D/T$ of the injected-target hypothesis ("screened": removed upstream before formation). L2/L3: Layer-2 exceedance and Layer-3 $D/T > 1$ epochs. Viol.: $e_H >$ HPL / $e_V >$ VPL at PL-valid epochs. Counts use the injection windows, except the full-sequence control and campaign rows. †One-epoch test in a separate run with otherwise identical settings.

TABLE V
LAYER-WISE OUTCOMES IN THE NOMINAL RUNS

| Layer | Outcome | Exp. A | Exp. B |
|---|---|---|---|
| 0 | FDE exclusions (meas./epochs) | 30/30 | 0/0 |
| 1 | Gate rejections (rejected/tested) | 6/20457 | 182/14938 |
| 2 | Exceedances with re-estimation | 1 | 3 |
| 3 | Risk computation available | 1311/1311 | 1242/1262 |
| 3 | SS alerts after remediation | 0 | 1 |
| 3 | PL unavailable (L2/SS/no PR) | 1/0/0 | 0/1/20 |
| 4 | SPP–FGO warning epochs | 21 | 42 |
| 4 | IMU-bias warning epochs | 0 | 0 |
| 5 | Normal/warning/alert | 1138/12/161 | 311/0/951 |

Counts are over evaluated navigation-output epochs. SS: solution separation; PR: pseudorange. Experiment B's three Layer-2 exceedances fall within its 20 no-pseudorange epochs, and the final PL-unavailable causes are mutually exclusive. Layer 5 gives ALERT precedence over WARNING, so Layer-4 warnings need not appear as final WARNING states.

Fig. 8(c) shows operational availability as a function of the alert limit. In Experiment A, horizontal availability rises from 3.5% at 10 m to 87.7% at 15 m and 98.7% at 20 m, so it is sensitive to the HAL within the range of loose safety-critical road applications; vertical availability is 88.6% at 25 m and 98.7% at 30 m. In Experiment B, availability at 60 m is 79.0% horizontally and 81.6% vertically, because 16.9% of epochs have no navigation solution or valid PL and the deep-urban PLs have heavy upper tails.

*2) Computational Efficiency:* Direct full-state inversion and the proposed rank update are compared on Experiment A under identical settings, both after FGO optimization. Total processing time includes graph construction, optimization, integrity monitoring, any Layer-2 re-estimation, and navigation output, but excludes initialization and standalone SPP/FDE. PL-module time includes posterior covariance extraction.

Table VI(a) reports reductions of 71.3% in all-hypothesis computation, 31.2% in PL-module time, and 24.5% in total processing time. Mean PL-module time decreases from 6.216 ms to 4.278 ms. Neither route changes the FGO problem, whose measured times differ only between independent replays. Both runs provide 1311 navigation outputs and 1310 valid PLs with matching solution-separation and PL-validity decisions; maximum HPL/VPL differences are 0.081/0.107 m, and one near-VAL epoch differs in vertical availability. Identical-input agreement is assessed below.

TABLE VI
RUNTIME COMPARISON (MS)

(a) Full-sequence processing, mean/p95

| Component | A: direct | A: proposed | B: proposed |
|---|---|---|---|
| Total processing | 9.173/15.978 | 6.928/12.452 | 5.952/12.706 |
| FGO optimization | 2.513/4.250 | 2.253/3.980 | 1.586/2.985 |
| PL module | 6.216/11.114 | 4.278/7.479 | 3.940/8.915 |
| All hypotheses | 2.190/4.264 | 0.629/1.088 | 0.512/1.745 |

(b) Matched-input timing per posterior

| | | Direct | | Proposed | | |
|---|---|---|---|---|---|---|
| Exp. | Scope | Mean ± SD | p95 | Mean ± SD | p95 | Reduction |
| A | (i) | $1.483 \pm 0.296$ | 1.809 | $0.407 \pm 0.085$ | 0.502 | 72.6% |
| A | (ii) | $1.604 \pm 0.334$ | 1.979 | $0.520 \pm 0.122$ | 0.658 | 67.6% |
| A | (iii) | $1.668 \pm 0.362$ | 2.065 | $0.580 \pm 0.136$ | 0.737 | 65.2% |
| B | (iii) | $0.249 \pm 0.193$ | 0.656 | $0.076 \pm 0.062$ | 0.207 | 69.6% |

(a) Separate 5 × full-sequence runs. (b) 13 A and 35 preselected B posteriors, all hypotheses per posterior. Cumulative scopes: (i) covariance and separation; (ii) plus nominal bias; (iii) plus detection and PL solving, including one nominal inversion for the direct method; FGO optimization and covariance extraction excluded. p95 of per-posterior medians; reductions compare means.

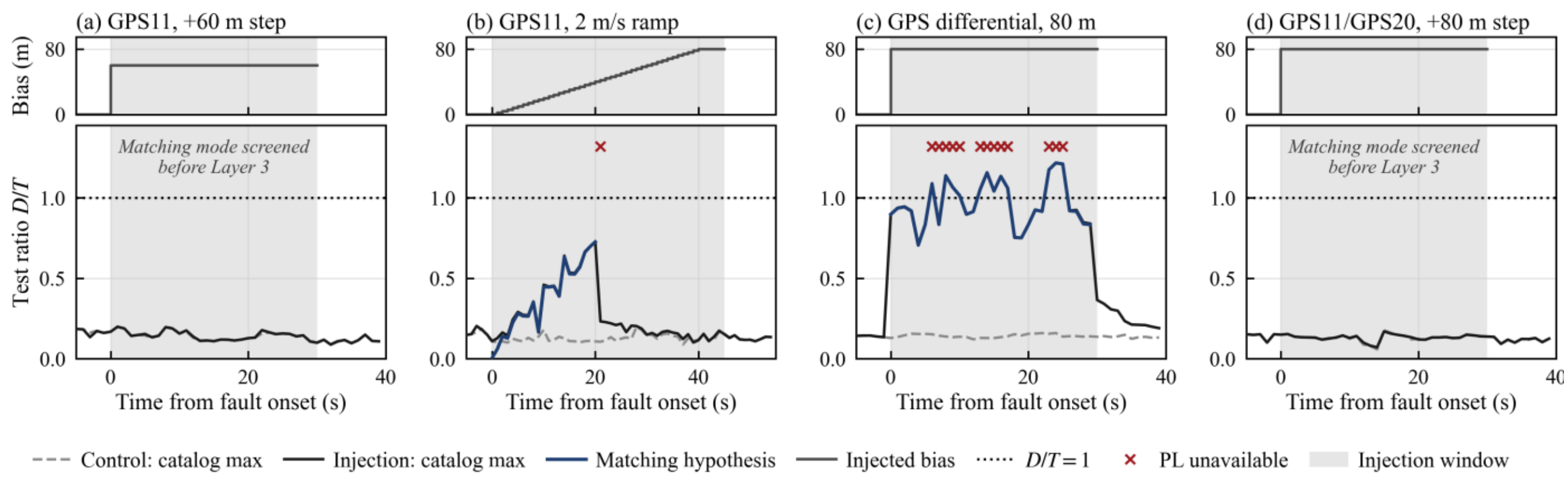

Fig. 7. Controlled pseudorange-fault response and Layer-2 remediation on Experiment A. Top: injected bias magnitude; bottom: solution-separation test ratio $D/T$. Gray dashed and black curves give the maximum $D/T$ over all hypotheses in the control and injection runs; the dark blue curve is the injected-target hypothesis when it survives screening. Red crosses mark PLs made unavailable by Layer 2 or a Layer-3 $D/T > 1$ event; shading marks the injection window.

Both formulations are evaluated on the same linearized FGO posteriors, with measurement data, fault hypotheses, risk parameters, and numerical tolerance held fixed (Table VI(b) and Table VII). Across 13 A and 950 B posteriors, computed HPL and VPL values agree at the 0.01 m numerical resolution, as do solution-separation alerts, PL-validity decisions, and alert-limit classifications, without regularization or eigenvalue truncation. In Experiment A, evaluating fault-dependent quantities and solving the PL equations requires 65.2% less time when the nominal posterior covariance is supplied; obtaining this covariance is included in online PL-module timing. The 35 preselected challenging Experiment-B posteriors are not representative of full-sequence runtime.

TABLE VII
FIXED-POSTERIOR NUMERICAL AGREEMENT

| Maximum discrepancy | Exp. A | Exp. B |
|---|---|---|
| Posteriors / hypothesis instances | 13/2155 | 950/100172 |
| Covariance element (m$^2$) | $4.832 \times 10^{-13}$ | $6.070 \times 10^{-7}$ |
| Relative cov. Frobenius norm | $2.517 \times 10^{-14}$ | $7.435 \times 10^{-12}$ |
| Position separation (m) | $8.993 \times 10^{-15}$ | $3.314 \times 10^{-8}$ |
| Retained-meas. bias sensitivity | $5.468 \times 10^{-15}$ | $6.028 \times 10^{-11}$ |
| Accumulated nominal bias (m) | $4.197 \times 10^{-14}$ | $1.423 \times 10^{-10}$ |
| SS detection threshold (m) | $5.007 \times 10^{-11}$ | $1.090 \times 10^{-8}$ |
| HPL / VPL (m) | 0 / 0 | 0 / 0 |

Direct inversion vs. rank update on identical posteriors (B: first 1200 s); instances are non-nominal hypotheses. HPL/VPL comparisons require passing solution-separation tests.

Fig. 9 resolves scope (i) by removal size and adds an intermediate full-state Woodbury route without selected output. The direct route costs 8.3–11.5 $\mu$s per hypothesis, dominated by the $23 \times 23$ factorization, whereas both Woodbury routes grow with $n_m$ through $\mathbf{\Sigma}_x \mathbf{H}_m^T$ and the $n_m \times n_m$ factorization. The speed-up of the proposed route therefore decreases from 5.0 for single-satellite modes to 2.1 for 12-measurement constellation or combined modes in Experiment A, and from 5.0 to 2.7 over the sizes observed in Experiment B, while exceeding 2 for every plotted size. Selected output reduces the full-state Woodbury time by 29.5% and 13.9% for $n_m = 1$ and 2 in Experiment A but by only 2.8–5.0% for $n_m \geq 5$, where the shared products dominate. Two-measurement modes account for 1474 of the 2155 Experiment-A hypothesis instances (68.4%); consequently, the aggregate reduction in Table VI(b) is governed mainly by small removal sets.

*3) Nonlinear Re-Optimization Assessment:* To assess the fixed-linearization approximation, the largest-separation single-satellite, constellation, and second-order modes are sampled at epochs 400, 800, and 1200 in a supplementary quadratic-loss run with Layer-2 remediation enabled. Each complete window is optimized with and without the hypothesized current-epoch pseudoranges, from the same online estimate, preserving the marginalization prior and its original linearization point. These offline solves neither replace navigation states nor enter the timing statistics.

TABLE VIII
AGREEMENT WITH NONLINEAR RE-OPTIMIZATION IN EXPERIMENT A

| Class | Factors removed | Conv. cases | 3-D sep. diff. (mm) | Cov. diff. ($10^{-5}$ m$^2$) |
|---|---|---|---|---|
| Single satellite | 1.00 | 3/3 | 0.0129/0.0249 | 2.524/4.698 |
| Constellation | 5.33 | 3/3 | 0.0333/0.0653 | 5.474/11.33 |
| Second order | 4.67 | 3/3 | 0.0494/0.1217 | 9.305/21.06 |

Mean/maximum over three hypotheses per class (epochs 400, 800, 1200) sharing the converged all-in-view control. Cov. diff.: Frobenius norm of the ENU position-covariance difference.

All three controls and nine fault-excluded solves satisfy a relative scaled-gradient tolerance of $10^{-6}$, taking 2–3 and 5–7 iterations, respectively; independent finite-difference gradients verify the controls. The all-in-view position changes by 0.073–0.183 mm. Rank predictions recomputed at these controls agree with the excluded-minus-control separation to within 0.122 mm and $2.106 \times 10^{-4}$ m$^2$ (Table VIII). These nine nominal-data samples assess local nonlinear agreement, not a universal approximation bound.

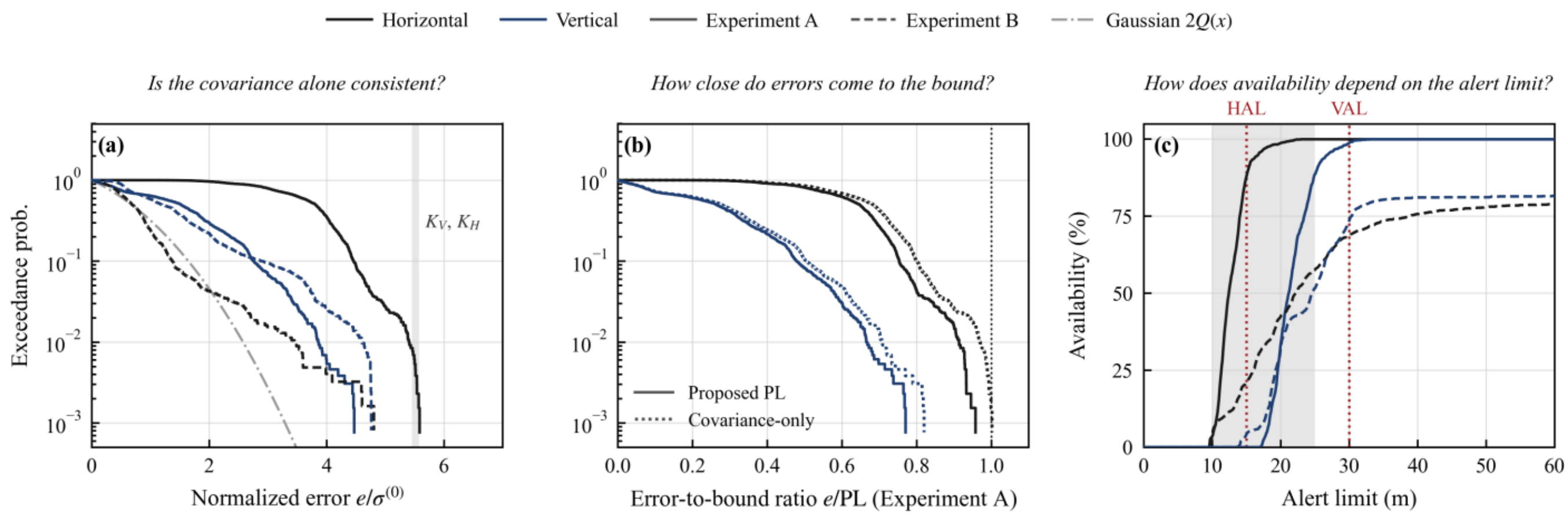

Fig. 8. Bound margins at PL-valid epochs. (a) Exceedance probability of the normalized error $e/\sigma^{(0)}$; curves above the Gaussian reference $2Q(x)$ indicate heavier tails, and the band marks $K_V$ to $K_H$. (b) Exceedance probability of the error-to-bound ratio in Experiment A for the proposed and covariance-only PLs. (c) Operational availability versus alert limit, including PL-unavailable and missing outputs; the band marks the 10–25-m HALs of loose safety-critical road applications [42].

*4) Fault-Hypothesis Coverage and Risk-Budget Ablation:* For the Experiment A ablation in Table IX, visible atomic events are held fixed while the monitored event types and maximum combination order are varied. Mutually exclusive probabilities and $P_{\text{not monitored}}$ are recomputed.

TABLE IX
FAULT-CATALOG RISK-BUDGET ABLATION

| Set | Ord. | Modes mean/max | $P_{\text{not monitored}}$ mean/max | Feasible/ evaluated |
|---|---|---|---|---|
| Experiment A | | | | |
| Fault-free | – | 1.00/1 | $4.558\times10^{-4}/4.799\times10^{-4}$ | 0/1311 |
| Sat. only | 2 | 132.31/172 | $2.999\times10^{-4}/3.000\times10^{-4}$ | 0/1311 |
| Const. only | 2 | 7.00/7 | $1.560\times10^{-4}/1.800\times10^{-4}$ | 0/1311 |
| Sat.+const. | 1 | 19.60/22 | $8.833\times10^{-8}/9.928\times10^{-8}$ | 1311/1311 |
| Sat.+const. | 2 | 169.49/214 | $9.687\times10^{-12}/1.180\times10^{-11}$ | 1311/1311 |
| Experiment B | | | | |
| Sat.+const. | 2 | 112.24/329 | $6.826\times10^{-12}/1.725\times10^{-11}$ | 1242/1262 |

Mode counts include the fault-free mode. Experiment B averages use 1242 computed catalogs; the other 20 outputs contain no current pseudorange.

A catalog with $P_{\text{not monitored}} \geq 10^{-7}$ cannot support a PL. Fault-free-only, satellite-only, and constellation-only catalogs are infeasible throughout the sequence. Both complete first- and second-order catalogs are feasible, but the first-order catalog consumes 88.3% of the total budget on average through unmonitored combinations. Second-order monitoring reduces that probability to approximately $10^{-11}$. This is a coverage/budget ablation, not a separate positioning run. Experiment B forms a feasible second-order catalog at 1242 of 1262 outputs, averaging 112.24 modes with a maximum of 329. Its 20 no-pseudorange outputs are excluded from this average and are not classified as risk-budget failures.

*5) Scope and Limitations:* Layer 3 evaluates protection levels for current-epoch GNSS pseudorange faults under a fixed linearization and a Gaussian error model. The exactness argument in Section IV-B4 relies on the removed factors acting only on the current state. A satellite fault that began earlier within the 15-s window also biases that satellite's historical factors, which are coupled through the other window states or absorbed into the marginalization prior; removing only the current-epoch factor then leaves part of the fault effect in the fault-excluded estimate, so the separation underestimates it. Such persistent faults must currently be caught by Layers 0–2 or by the growing current-epoch separation; the 2-m/s ramp in Section V-B2 produced no bound violation, but a single test does not establish protection in general. A complete treatment would extend each hypothesis to all window factors of the affected satellite, using the joint window covariance and a bound on the fault contribution retained in marginalization priors, analogous to sequential Kalman-filter integrity monitoring [26], [27]. Doppler and IMU faults are outside the Layer-3 hypothesis set.

The PL is also conditioned on the measurement selection of Layers 0–2. Baseline ARAIM accounts for exclusion by adding the probability of wrong exclusion to the integrity risk of the post-exclusion PL [9], [20]; this term is not included here. Every retained satellite remains covered by the single- and second-order hypotheses, so a fault that survives a wrong exclusion is still monitored, but the selection itself is data dependent. In Experiment A, the 30 epochs with a Layer-0 exclusion all retained valid PLs, with maximum error-to-HPL and error-to-VPL ratios of 0.76 and 0.37, and the maximum error-to-HPL ratio at the following epochs was also 0.76. Including the exclusion risk of Layers 0 and 2 in the risk allocation remains future work.

The false-alert interpretation of the separation threshold assumes a zero-mean continuity model and a fault-free historical prior; the effects of bounded mean biases and adaptive screening and re-estimation on it require further analysis. The risk budget applies to each PL evaluation rather than per hour, and the experiments do not establish the specified rare-event integrity risk (Section V-D1); the posterior uncertainty and nominal-bias models require validation on independent data, for example through Gaussian overbounding of sample error distributions [35] and covariance overbounding under time-correlated noise [45].

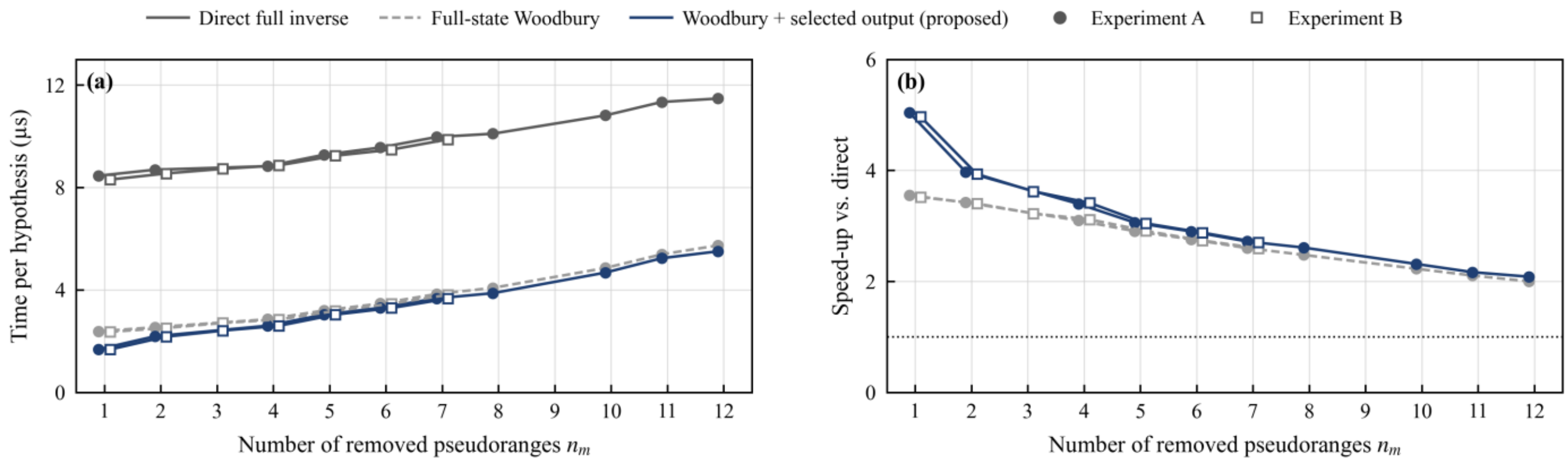

Fig. 9. Per-hypothesis computation time for scope (i) of Table VI(b) versus the number $n_m$ of removed pseudoranges, on the shared posteriors (13 in A, 35 in B): (a) mean time per hypothesis instance; (b) speed-up relative to direct full-state inversion. The full-state Woodbury route evaluates the Woodbury covariance without selected output. Sizes with fewer than five instances are omitted, and Experiment B markers are offset horizontally for visibility.

## VI. Conclusion

This paper formulated ARAIM solution separation, fault-dependent position covariance, and nominal-bias propagation with integrity-risk allocation on the posterior of tightly coupled GNSS/IMU FGO, exactly for the linearized fixed-lag problem because current-epoch pseudorange factors act only on the current state. Woodbury removal with selected-output evaluation preserves full-state correlations without full-state inversion or graph re-optimization per hypothesis, and the surrounding layers provide measurement screening, residual-based re-estimation, consistency checks, and availability decisions.

Both road experiments achieved 100% horizontal and vertical empirical bounding over PL-valid epochs, with a maximum error-to-PL ratio of 0.96. Normalized errors revealed a slowly varying common-mode ranging offset that the posterior covariance does not represent, so a covariance-only bound retained no resolvable margin, whereas the nominal-bias and fault-hypothesis terms increased the mean PLs by 4.0–11.7%. With only 9–44 effectively independent samples per channel, the experiments are consistency checks rather than validation of the integrity budget. On identical Experiment A posteriors, the proposed method reduced fault-hypothesis evaluation and PL solving time by 65.2% relative to direct full-state inversion, with matching integrity decisions and HPL/VPL values, and the per-hypothesis speed-up remained above 2 for removal sets of up to 12 pseudoranges. Fault injections demonstrated hypothesis responses and fail-safe PL invalidation while navigation continued. Empirical bounding did not, however, ensure availability in the deep-urban sequence. Further work will address persistent faults absorbed into historical priors through window-level satellite hypotheses, the exclusion risk of the screening layers, and validation of uncertainty and bias models on larger independent data sets.